*MetaClimage: A novel database of visual metaphors related to Climate Change, with costs and benefits analysis*

*Biagio Scalingi[1,*], Chiara Barattieri di San Pietro[1], Paolo Canal[1], Valentina Bambini[1]*

[1] *Laboratory of Neurolinguistics and Experimental Pragmatics (NEPLab), University School for Advanced Studies IUSS, Pavia, Italy.*

[*] *Corresponding author: Biagio Scalingi, Laboratory of Neurolinguistics and Experimental Pragmatics (NEP), University School for Advanced Studies IUSS, Pavia, Italy, Piazza della Vittoria 15, 27100 Pavia, Italy; email:* biagio.scalingi@iusspavia.it

*Abstract*

Visual metaphors of climate change (e.g., melting glaciers depicted as a melting ice grenade) are regarded as valuable tools for addressing the complexity of environmental challenges. However, few studies have examined their impact on communication, also due to scattered availability of material. Here, we present a novel database of Metaphors of Climate Change in Images (*MetaClimage*) https://doi.org/10.5281/zenodo.15861012, paired with literal images and enriched with human ratings. For each image, we collected values of difficulty, efficacy, artistic quality, and emotional arousal from human rating, as well as number of tags generated by participants to summarize the message. Semantic and emotion variables were further derived from the tags via Natural Language Processing. Visual metaphors were rated as more difficult to understand, yet more aesthetically pleasant than literal images, but did not differ in efficacy and arousal. The latter for visual metaphors, however, was higher in participants with higher Need For Cognition. Furthermore, visual metaphors received more tags, often referring to entities not depicted in the image, and elicited words with more positive valence and greater dominance than literal images. These results evidence the greater cognitive load of visual metaphors, which nevertheless might induce positive effects such as deeper cognitive elaboration and abstraction compared to literal stimuli. Furthermore, while they are not deemed as more effective and arousing, visual metaphors seem to generate superior aesthetic appreciation and a more positively valenced experience. Overall, this study contributes to understanding the impact of visual metaphors of climate change both by offering a database for future research and by elucidating a cost-benefit trade-off to take into account when shaping environmental communication.

# 1. Introduction

Climate change (CC) refers to significant alterations in the global surface temperature compared to the years 1850-1900. The consequences of CC impact several aspects of the world, from ecosystems and agriculture (Eigenbrode et al., 2022) to the physical and biogeochemical conditions of the oceans (Wåhlström et al., 2022), and human health (Ebi et al., 2018). Hence, the topic of CC, initially limited to discussions inside small groups of experts, has now become a matter of concern for millions of individuals, especially for matters such as greenhouse gases, renewable energy, and decarbonization (Boykoff, 2008; Effrosynidis et al., 2022). However, despite the consensus in the scientific community, there still is a critical need to raise awareness in society at large regarding CC causes and effects, as well as its possible solutions, which makes CC communication pivotal at many levels.

Communicating complex subjects such as CC must go beyond the mere amount of information delivered and raises the issue of how to convey the message to non-experts (Nerlich et al., 2010; Smillie & Scharfbillig, 2024; Yusuf & St.John III, 2021). For this purpose, visual communication proved to be helpful. For instance, the use of pictures to discuss CC can help people perceive the phenomenon as more immediate in both spatial and temporal terms (Duan et al., 2019). Similar effects were also observed in communication based on verbal metaphors (Turner & Littlemore, 2023). When people are presented with a text about CC using a war metaphorical frame (i.e., "the war against climate change"), they consider CC as a more urgent problem and are more willing to be pro-environmental compared to participants who are presented with a text about CC without a metaphorical frame (i.e., "the issue of climate change"; Flusberg et al., 2017). The communicative potential of metaphors and images suggests the combination of the two, i.e., via visual metaphors, might be particularly effective in delivering CC messages and raising awareness. Recent years have witnessed a growing interest in visual metaphors to convey information about CC (Augé, 2022; Cozen, 2013; Platonova, 2019).

Visual metaphors are pictures with a perceptual incongruity between two core elements (Forceville, 2014). In a Relevance theory framework (Sperber & Wilson, 1995), the resolution of perceptual incongruities involves both decoding and inferences, as visual metaphors prompt the viewers to retrieve visual elements from memory and construct metaphorical *ad hoc* concepts through broadening and narrowing processes (Yus, 2009), reconciling the anomaly with a metaphorical meaning. Take the example in Figure 1A, where melting glaciers are depicted as a melting ice grenade prompting the viewer to broaden the concept GRENADE in order to include ice melting as a dangerous phenomenon.

Visual metaphors have been extensively studied in the advertising literature due to their frequent use in commercials (Phillips & McQuarrie, 2004; Toncar & Munch, 2001; Ventalon et al., 2020). A specific property of visual metaphors that emerges from the empirical literature is that they are often more costly than literal messages (Bambini et al., 2024; Ojha et al., 2019). In particular, the use of visual metaphors must consider a delicate trade-off between costs and benefits: the more cognitive effort the addressee needs to invest to retrieve the message of the image, the less relevant (Sperber & Wilson, 1995) and the less appreciated that message becomes (Van Mulken et al., 2014). More generally, moderately difficult metaphors are accompanied by a number of communicative benefits, such as artistic appreciation (Bolognesi & Werkmann Horvat, 2022; McQuarrie & Mick, 2003) and arousal (Forceville, 2008). Another aspect playing a role is individual differences in appreciating visual metaphors. Given the complexity of these images, a dimension that has been particularly considered in this respect is the Need for Cognition (NFC), which refers to a person's tendency to engage in and benefit from cognitive effort (Cacioppo & Petty, 1982). In the case of metaphors, higher NFC was associated with a better appreciation of the message as well as higher subjective comprehension (Chang & Yen, 2013; Mohanty & Ratneshwar, 2014; Phillips & McQuarrie, 2004).

The use of visual metaphors in CC communication is not rare (Platonova, 2019). For instance, CC communication frequently employs the metaphorical frames of war, violence, and destruction characterized by negative valence, as in the case of melting grenade-glaciers in Figure 1A (Hidalgo-Downing & O'Dowd, 2023), a dolphin squeezed as a wet T-shirt in Figure 1B, or car exhaust fumes taking the shape of the Amazon Rainforest in Figure 1C. Other examples were described by Pérez-Sobrino (2016) under the label of 'shockvertisements', i.e., advertisements that combine unusual elements in strange scenarios to call for attention towards a certain issue, such as, for instance, a fox wearing a naked woman around its neck like a fur coat (Figure 1D).

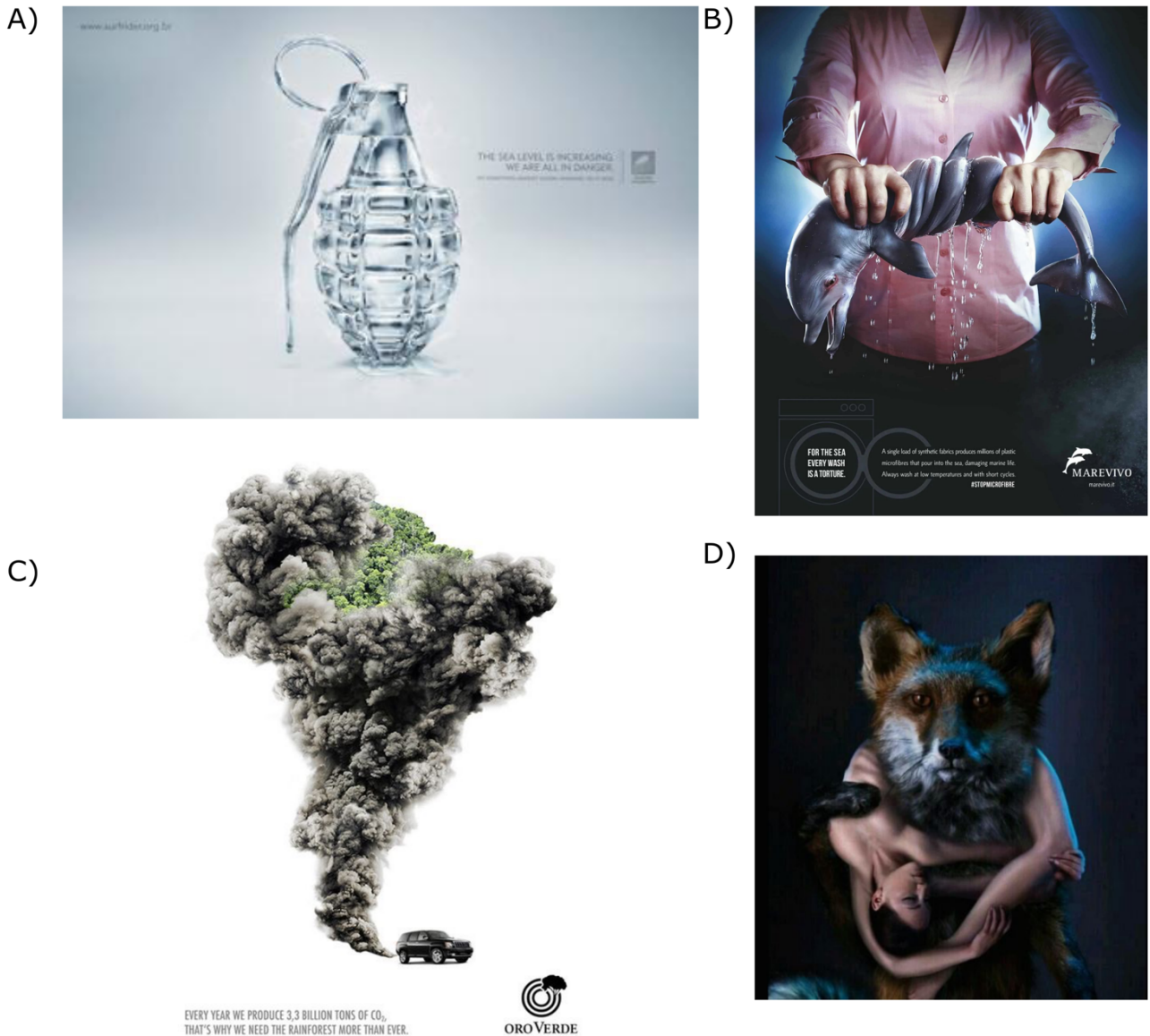

**Figure 1**. Examples of visual metaphors of Climate Change from advertising campaigns. The image in A comes from the "Grenade, Bomb" campaign by Surfrider Foundation and is shown in lower resolution to allow dissemination for scientific purposes. The image in B is part of the "Stop Microfibre" campaign by Marevivo, created by Daniele Piergiovanni and Rosario Oliva, and is licensed under CC BY-NC-ND 4.0 (https://www.behance.net/gallery/73934663/MAREVIVO). The image in C comes from the "Oro Verde – Smoke" campaign by Oro Verde, created by Patrick Ackmann, and is licensed under CC BY-NC-ND 4.0 (https://www.behance.net/gallery/3464443/Oro-Verde-Smoke). The image in D is part of the "Against Fur" by The Association Against Animal Abuse and is shown in lower resolution to allow dissemination for scientific purposes.

Prior literature (Hidalgo-Downing & O'Dowd, 2023) concentrated on the descriptive level (i.e., identifying the domains of the metaphors), while, to the best of our knowledge, no studies investigated the processing costs and the communicative benefits associated with visual metaphors in CC context.

One of the reasons behind the current state of the art might be the absence of corpora and databases of CC images, which makes it difficult to identify and choose the most appropriate material to use. Even though CC visual datasets exist (Lehman et al., 2019; Magalhães et al., 2018), they include only literal stimuli. Also, most of the existing databases do not provide cognitive and psycholinguistic metrics, which may be pivotal in shaping investigations on CC communication (Thieleking et al., 2020).

The primary aim of our study was to create a database of CC pictures, including visual metaphors and literal stimuli, enriched with relevant measures derived via subjective ratings. Secondly, we aimed to provide information on processing costs and communicative benefits associated with visual metaphors, by analyzing the rating variables and by complementing the study with a tag generation task useful to ascertain further aspects of comprehension. To pursue these aims, we started by building a database of visual metaphors of CC, gathering pictures collected in content creator networks, already existing databases and stock sites. In order to focus on the dramatic consequences of CC, we decided to include only metaphorical pictures with a negative valence; also, we decided to include in the database also literal stimuli, i.e., realistic pictures of CC, in order to provide a benchmark with respect to which specific effects of metaphorical conceptualization could be considered. We named the final database *Metaphors of Climate Change in Images* (*MetaClimage*). Furthermore, to derive rating values, we recruited a sample of participants who were asked to judge difficulty, efficacy, artistic quality, and arousal of each image, in addition to being measured for their NFC. Participants were also asked to complete a tag generation task (Bolognesi et al., 2019), requesting to summarize the message of each picture through tags and deriving measures such as number of tags, number of words in the tags, ratio between non-depicted and depicted tags. To the tags, we then applied Natural Language Processing (NLP) tools in order to extract further semantic and emotion variables (i.e., concreteness, imageability, semantic similarity, valence, word arousal, and dominance). All these values were included as additional measures to enrich the MetaClimage database and to conduct the cost and benefit analysis.

Overall, we hypothesized that visual metaphors would be associated with greater costs yet more benefits compared to literal stimuli. Specifically, we expected visual metaphors to be rated as more difficult to understand (van Mulken et al., 2014) but also as more effective (Phillips & McQuarrie, 2002), with higher artistic quality (Bolognesi & Werkmann Horvat, 2022), and greater arousal (Forceville, 2008) compared to literal stimuli. We also expected that people with higher NFC would pay less effort and greater appreciation when presented with visual metaphors (Mohanty & Ratneshwar, 2015; Phillips & McQuarrie, 2004; Chang & Yen, 2013). As for the tag generation task, visual metaphors were expected to trigger more tags, with more words and increased reference non-depicted objects (Šorm & Steen, 2013), which may further indicate the greater effort in cognitive elaboration (Bolognesi & Vernillo, 2019; Ventalon et al., 2020). Lastly, we expected that the NLP variables extracted from the tags would mirror the ratings and reflect the greater elaboration costs as well as the greater benefits of visual metaphors. Hence, concerning semantic variables, we expected, for instance, more abstract words and lower semantic similarity in the tags for visual metaphors, and, concerning emotional variables, we expected that more emotionally charged words (i.e., lower valence, higher word arousal, lower dominance) would be used for describing visual metaphors compared to literal images.

**2. Materials and methods**

*2.1 Development of the MetaClimage database*

The procedure we followed to build the MetaClimage database is summarized in Figure 2. We first searched for visual metaphors on web engines using the keywords "climate change metaphor", "deforestation metaphor", "drought", and "plastic pollution". On the retrieved images, the first selection aimed at excluding those pictures not matching the following criteria: (i) dealing with environmental issues and (ii) having a negative connotation. For instance, a picture showing a politician talking about CC would be discarded because not clearly representing any CC issue nor having a negative valence. Each image was then evaluated according to VISMIP criteria (Šorm & Steen, 2018): a) identifying the general understanding of the image, b) describing the picture elements, c) identifying the incongruous element generating the perceptual incongruity, d) searching for an element able to restore the incongruity; e) assessing whether there could be established a cross-domain comparison between the elements c and d. Literal stimuli were searched for partly on web engines, using the same keywords used to search for visual metaphors but removing "metaphor" from the string, partly from the Affective Climate Images database (Lehman et al., 2019). For literal items too, we excluded those pictures not dealing with environmental issues and/or nor having negative connotation.

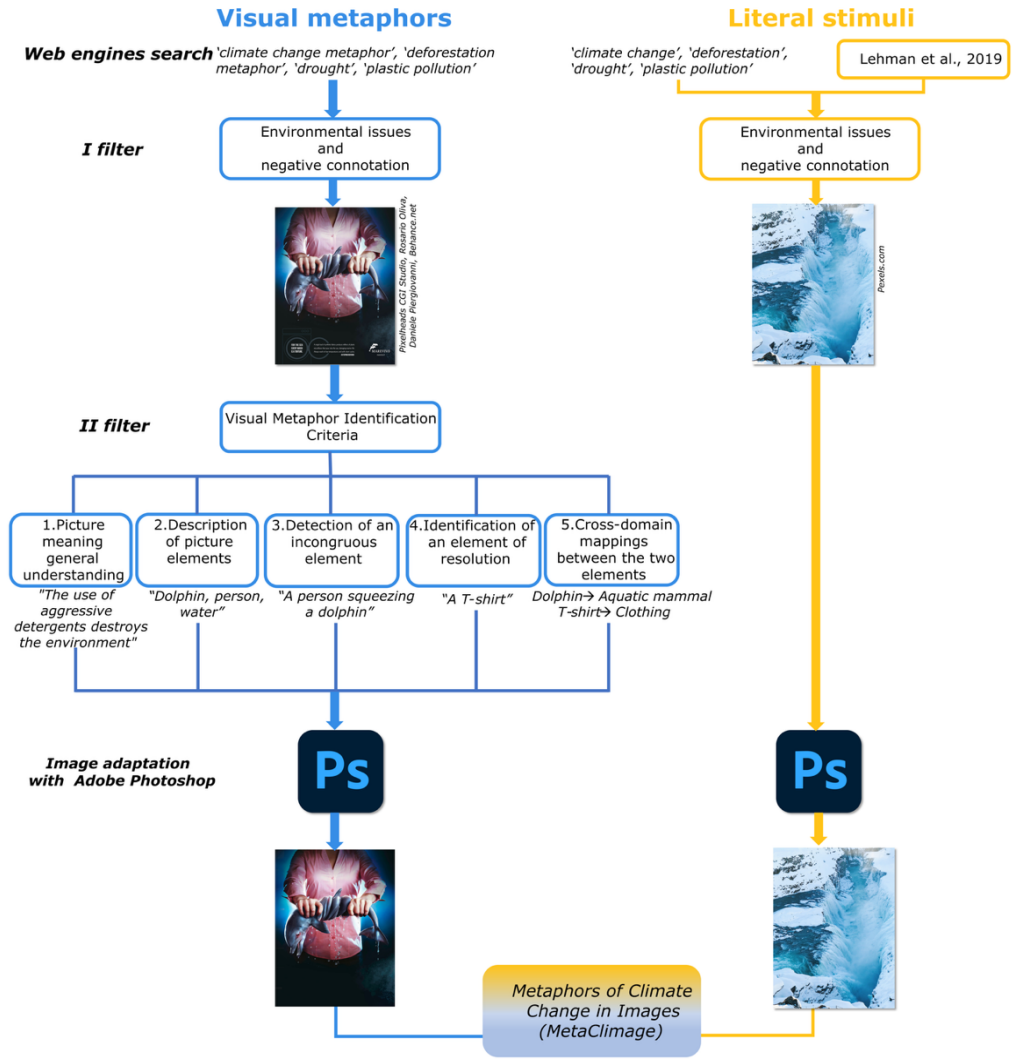

**Figure 2.** Workflow of MetaClimage construction. Pipeline for visual metaphors and literal stimuli selection.

The final database is made up of 100 CC-related images, 50 metaphorical and 50 literal. Of the 50 visual metaphors in the database, 16 (32%) pictures were taken from NGOs campaigns and sites (e.g., AdsoftheWorld - adsoftheworld.com, Adforum - adforum.com, and Surfrider - surfrider.org); 16 (32%) from the Behance content creators network (behance.net); 7 (14%) from image stock sites (i.e., AdobeStock - stock.adobe.com and Flickr - flickr.com); d) 9 (18%) from forums and social platforms (i.e., DeviantArt - deviantart.com, Pinterest - pinterest.com); and 2 (4%) from sites of designers (e.g., Federico Pestilli). Of the 50 literal stimuli, N = 17 pictures (34%) were taken from Lehman and colleagues' database (2019), 29 (58%) from stock sites (i.e., Pexels - pexels.com and Pixabay - pixabay.com), 3 (6%) from websites that did not report the author's name; and 1 (2%) from sites of private authors. In the whole database 30 (30%) images illustrated plastic, 20 (20%) deforestation, 20 (20%) animals, 16 (16%) pollution, 7 (7%) glaciers, and 7 (7%) drought (see Figure 3).

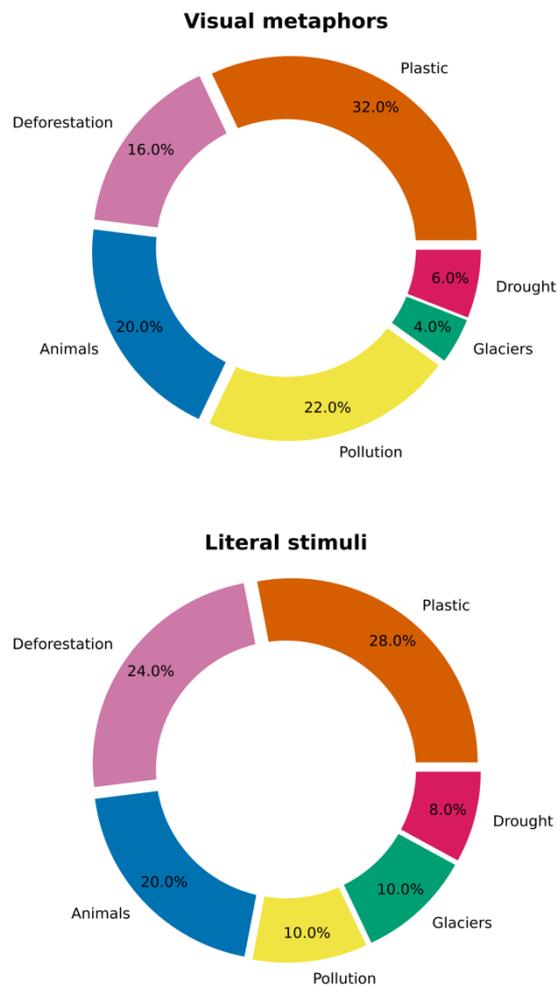

**Figure 3.** Distribution of environmental issues for visual metaphors and literal stimuli in the MetaClimage database.

Following Phillips and McQuarrie's (2004) taxonomy, the visual metaphors were distributed as follows: 62% (N = 31) replacements, where one image stands in for another absent element—as in met_29, where a dolphin is squeezed like a wet T-shirt (see Figure 1B), 36% (N = 18) fusions, where two elements are merged into a single image—as in met_27, where car exhaust smoke takes the shape of the Amazon Rainforest and almost entirely covers it (see Figure 1C), and 2% (N = 1) juxtapositions, in which two elements are placed side by side—such as in met_40, where a tree is shown next to a cut finger.

Writings and captions were manually removed from pictures using Adobe Photoshop CC 2019. After adaptation, pictures from the two datasets were compared based on three structural features: luminance (i.e., the overall brightness of the image), contrast (i.e., the variation in pixel values around the mean luminance), and entropy (i.e., the richness of details within the image). Solely for the purposes of measuring luminance, contrast, and entropy, each image was converted to grayscale.

Luminance was calculated as the mean pixel value, contrast as the standard deviation of pixel values, and entropy was computed by generating a histogram of intensity distribution using $N = 128$ bins and measuring the randomness in the pixel intensity probability distribution. Metaphorical and the literal dataset did not differ in terms of luminance (0.55 for visual metaphors and 0.50 for literal stimuli; $p = 0.16$) nor contrast (0.20 for visual metaphors and 0.22 for literal stimuli; $p = 0.07$), but a significant difference emerged for entropy ($p < 0.001$), which was lower in visual metaphors (6.15 vs 6.68), suggesting that they were more uniform and contained fewer elements.

*2.2 Rating and tag generation study*

2.2.1 Participants

Seventy-six participants (N = 76, 45F, 60.81%) were enrolled in the study. The mean age was 27.08 years (SD = 4.74), and the mean education was 16.89 years (SD = 2.15). Participants were recruited through acquaintances, referrals, and interpersonal networks of the researchers. The Ethics Committee of the Department of Brain and Behavioral Sciences of the University of Pavia approved the protocol (n.121/23). Participants signed the informed consent before the study. The study was conducted following the Declaration of Helsinki.

2.2.2 Procedure and assessment

Participants were sent a link to the LimeSurvey® platform. After providing informed consent and demographic information (i.e., age, gender, and education level) through a web form, participants were asked to complete the short version of the Need for Cognition scale (Cacioppo et al., 1984), which measures an individual's tendency to engage in and enjoy effortful cognitive activities. The scale consists of 18 items rated on a 5-point Likert scale, ranging from "Totally disagree" to "Totally agree", with a total score ranging from -36 to 36. Then participants were presented with the image rating and tag generation task.

2.2.3 Image rating and tag generation task

Each participant was assigned to one of three lists, each containing 34 or 33 stimuli, with a similar proportion of visual metaphors and literal stimuli. Participants were asked to rate each image's difficulty ("How difficult is it to understand the message conveyed by the image?"), efficacy ("How effective is the image in communicating a message about environmental issues?"), artistic quality ("How do you rate the artistic quality of the image?"), and arousal ("How much does the image excite you?") on a 7-point Likert scale. Lastly, for each image, participants were required to complete a tag generation task (Bolognesi et al., 2019), which consisted in summarizing the picture message using keywords ("Describe the message this image is attempting to convey using keywords").

*2.3 Extraction of variables from the tags*

In order to further investigate the costs and benefits of visual metaphors, we extracted several variables from the tags produced in the tag generation task. First of all, for each image, we derived tag number, tag word number, and the non-depicted/depicted ratio. The latter refers to the ratio between the number of tags classified as referring to something not depicted (i.e., non explicitly visible in the image but that could be derived through inferential procedure) and the number of tags classified as referring to something depicted (i.e., explicitly visible in the image). The classification was conducted manually, following Bolognesi and colleagues (2019). For instance, for item lit_72 (a polar bear jumping from one glacier to another), a tag such as "ice" was classified as depicted, whereas a tag such as "extinction risk" tag was classified as non-depicted. The non-depicted/depicted ratio was high when participants generated many non-depicted tags and low when participants generated many depicted tags.

Another set of variables was extracted from tags via NLP. Tags underwent pre-processing, consisting of Italian stopwords removal by means of the Python *NLTK* package (Bird et al., 2009), tokenization and lemmatization by means of the *Spacy* package (Honnibal & Montani, 2017). Each resulting lemma was characterized along the emotional dimension (valence, arousal, dominance) and for semantic features (concreteness, imageability) extracted from the *MEmoLon* database (Buechel et al., 2020) and the *MEGAHR* database (Ljubešić et al., 2018), respectively. For each image, a mean value was calculated for each variable. Then, word embeddings for each word in every tag were extracted using *fasttext* (Bojanowski et al., 2017) to compute semantic similarity. A similarity matrix was generated, reporting the semantic similarity values for every pair of tag word vector. Finally, we calculated the grand average of semantic similarity describing the overall similarity of a matrix in which the row and column vectors represent words generated for each image.

*2.4 Statistical analysis*

Firstly, we tested raters' reliability by computing the Intraclass Correlation Coefficient (ICC), selecting a two-way model testing agreement on the average score with the *irr* package (Gamer et al., 2022).

Secondly, we assessed the differences between literal and metaphorical images in rating variables (difficulty, efficacy, artistic quality, and arousal), tag variables (tag number, word number, non-depicted/depicted ratio), semantic (concreteness, imageability, and semantic similarity) and emotion variables (valence, word arousal, and dominance). Differences between conditions were assessed with Student-*t* tests for independent samples when normality was not rejected or with a non-parametric test (Mann–Whitney *U* test) otherwise, corrected for multiple comparisons using False Discovery Rate (FDR).

For each condition (metaphorical and literal), we performed correlation analysis between rating variables and, separately, between tag variables. Furthermore, we correlated rating and tag variables with semantic and emotion variables. The strength of the correlations was then compared between types of image (metaphorical vs. literal), after performing Fisher's z-transformation on the correlation coefficients.

Finally, participants' level of NFC was calculated and correlated with rating, tag, semantic, and emotion variables.

Statistical analyses were performed using the *scipy.stats* package (Virtanen et al., 2020) implemented in Python (version 3.10.12).

## 3 Results

*3.1 Results of the image rating task*

Table 1 and Figure 4 display the results of the image rating task. ICC showed moderate agreement between the raters for efficacy, difficulty, artistic quality, and arousal ($ICC_{Efficacy}$= 0.66, 95% CI [0.53, 0.75], $p < 0.001$; $ICC_{Difficulty}$= 0.69, 95% CI [0.6, 0.77], $p < 0.001$; $ICC_{Artistic\ Quality}$= 0.61, 95% CI [0.50-0.71], $p < 0.001$; $ICC_{Arousal}$= 0.57, 95% CI [0.46, 0.67], $p < 0.001$).

*Difficulty*: Visual metaphors were judged of medium-low difficulty (M = 3.20), with scores ranging from 1.88 (met_50: a goldfish embedded inside a plastic cup) to 5.19 (met_48: a piece of meat shaped like a footprint). Literal stimuli were judged as having low difficulty (M = 2.84), with scores ranging from 1.33 (lit_93: a set of plastics in a waterway) to 4.92 (lit_81: melting glaciers). The comparison between conditions was significant ($p = 0.02$), indicating that the interpretation of the message of visual metaphors was judged as more complex.

*Efficacy*: Visual metaphors were considered as highly effective (M = 5.03), ranging from 3.26 (met_42: a smoking cloche with the world drawn on) to 6.24 (met_04: a group of burning trees arranged in a lung-like shape). The efficacy of literal stimuli was also high (M = 5.02), with scores ranging from 1.33 (lit_93: a set of plastics in a waterway) to 4.92 (lit_81: melting glaciers). The comparison was not significant ($p = 0.96$), indicating that the two types of images were considered as equally effective in conveying CC-messages.

*Artistic quality*: The artistic quality of visual metaphors was judged as high (M = 5.33), ranging from 4.22 (met_48: a piece of meat shaped like a footprint) to 6.32 (met_04: a group of burning trees arranged in a lung-like shape). The artistic quality of literal stimuli was medium-high (M = 4.68), ranging from 3.81 (lit_59: industrial smokestacks releasing smoke) to 6.16 (lit_51: a fish trapped in a plastic glove). The comparison was significant ($p < 0.001$), indicating that visual metaphors were considered as more aesthetically pleasing than literal stimuli.

*Arousal*: The arousal ratings for metaphors were high (M = 5.14) and ranged from 4.04 (met_16: a green plastic veil reminiscent of the forest landscape) to 6.04 (met_29: a dolphin squeezed like a wet T-shirt). Arousal value for literal stimuli was also high (M = 5.12), ranging from 4.37 (lit_52: a broken plastic bucket on a beach) to 6.12 (lit_82: a turtle with its shell shattered in the sea). The comparison was not significant ($p = 0.83$), indicating that metaphors and literal stimuli were equally impactful at the emotional level.

**Table 1.** Results of the rating and tag generation task for visual metaphors and literal stimuli included in MetaClimage

| Variables | Visual Metaphors M (SD, Median) | Literal stimuli M (SD; Median) | 95% CI | Statistics | p values |
|---|---|---|---|---|---|
| Efficacy | 5.03 (0.76, 5.16) | 5.02 (0.66, 5.06) | -0.27 - 0.29 | $t(98) = 0.06$ | 0.96 |
| Difficulty | 3.20 (0.79, 3.04) | 2.84 (0.69, 2.87) | 0.07 - 0.66 | $t(98) = 2.45$ | 0.02 |
| Artistic Quality | 5.33 (0.49, 5.23) | 4.68 (0.48, 4.62) | 0.46 - 0.85 | $t(98) = 6.73$ | <0.001 |
| Arousal | 5.14 (0.47, 5.14) | 5.12 (0.51, 5.08) | -0.17 - 0.22 | $t(98) = 0.21$ | 0.83 |
| Tag number | 2.07 (0.40, 2.00) | 1.84 (0.20, 1.80) | 0.06 - 0.28 | $U = 1715.5$ | 0.001 |
| Word number | 1.51 (0.29, 1.47) | 1.44 (0.15, 1.45) | -0.24 - 0.16 | $t(98) = 1.48$ | 0.14 |
| Non-Depicted/Depicted | 5.00 (6.33, 2.96) | 1.33 (1.20, 1.09) | 1.27 - 2.86 | $U = 2176.5$ | < 0.001 |
| Concreteness | 3.35 (0.18, 3.34) | 3.30 (0.17, 3.29) | 3.35 - 3.30 | $t(98) = 1.56$ | 0.12 |
| Imageability | 3.67 (0.13, 3.65) | 3.63 (0.12, 3.61) | -0.01 - 0.09 | $t(98) = 1.68$ | 0.10 |
| Semantic similarity | 0.37 (0.04, 0.38) | 0.41 (0.04, 0.41) | -0.06 - 0.02 | $t(98) = -4.72$ | < 0.001 |
| Valence | 4.52 (0.25, 4.53) | 4.26 (0.31, 4.22) | 4.52 - 4.26 | $t(98) = 4.67$ | < 0.001 |
| Word arousal | 4.34 (0.15, 4.31) | 4.40 (0.18, 4.39) | -0.13 - 0.01 | $t(98) = -1.82$ | 0.07 |
| Dominance | 4.70 (0.17, 4.69) | 4.53 (0.18, 4.51) | 0.11 - 0.25 | $t(98) = 5.04$ | < 0.001 |

*Note:* Average values and standard deviations (in parentheses) are provided. The last column displays the contrast between the two conditions (via independent samples *t*-tests or Mann–Whitney *U* tests).

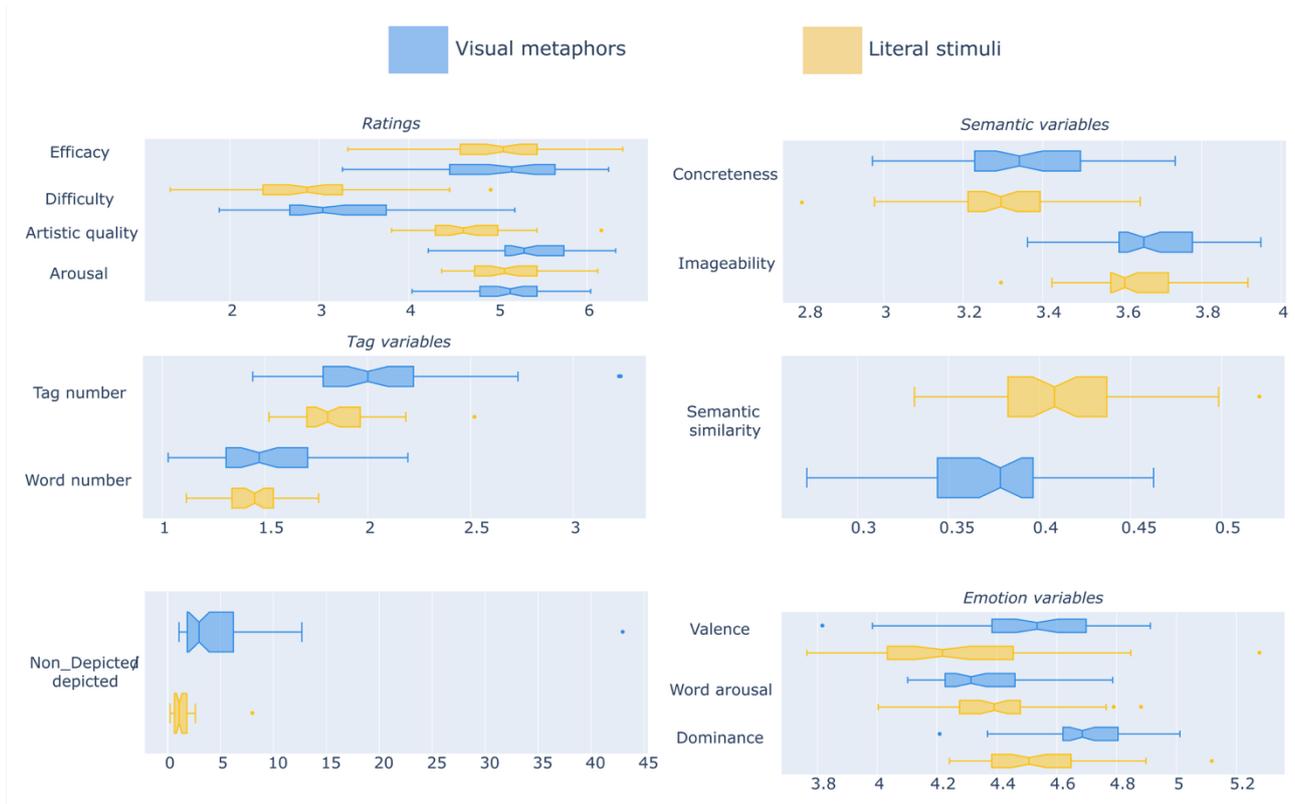

**Figure 4.** Results from the rating and the tag generation task. Comparisons between visual metaphors and literal stimuli across ratings, tag, semantic, and emotion variables.

*3.2 Results of the tag generation task, semantic and emotion variables*
Table 1 and Figure 4 display the results of the tag generation task. For visual metaphors, the mean number of tags was higher (2.07) than for literal stimuli (1.84; $p = 0.001$), with also a greater non-depicted/depicted ratio (5.00 vs. 1.33; $p < 0.001$). The mean number of words per tag did not differ between conditions. To exemplify this pattern, consider the comparison between met_37, representing a knife as a bulldozer clearing a forest that turns into chocolate cream, which received on average 2.15 tags, with many tags describing concepts not explicitly shown in the image, such as "logging", "deforestation", and "consumerism" (mean non-depicted/depicted ratio 5.44), and lit_59, representing a countryside with a factory emitting fumes, which received on average 1.70 tags, with the most common tags including "smog", "smoke", and "factory", all referring to something clearly visible in the picture (mean non-depicted/depicted ratio 0.24).

Concerning semantic variables, the tags of visual metaphors and literal stimuli did not differ in terms of concreteness and imageability ($p = 0.12$ and $p = 0.10$, respectively), suggesting that the two types of items elicited the production of words that describe similar concepts in terms of bodily experience and their ease of generating a mental image. Yet, a significant difference emerged in semantic similarity, with tags for visual metaphors displaying lower values than literal items, i.e., encompassing semantically more distant concepts ($p < 0.001$). As an example, consider the comparison between met_34, depicting a roll of paper with the world drawn, which generated tags referring to relatively distant concepts, such as "consumption", "injured globe", "logging", "medicated patch" (mean semantic similarity 0.33), and lit_77, depicting a dry ear of wheat, which generated semantically closer concepts such as "climate change", "environmental disaster", "climate crisis", "environmental pollution" (mean semantic similarity 0.46).

Concerning emotion variables, tags for visual metaphors were characterized by higher valence ($p < 0.001$), i.e., they included words describing more pleasant situations compared to tags for literal stimuli which included more words describing unpleasant situations. No difference was observed in word arousal between the two conditions ($p = 0.07$). However, tags for visual metaphors displayed significantly higher dominance ($p < 0.001$), i.e., they included more words referring to in-control situations compared to literal stimuli, in which words referred more to out-of-control situations. For instance, met_09, depicting a glass of water with melting ice and polar bears on top, generate tags (e.g., "water","glass", and "melting") with a mean valence of 4.53, mean dominance of 4.67, and mean arousal of 4.26. In contrast, lit_64, depicting people on melted glaciers, generated tags (e.g., "death" and "danger") with a mean valence of 4.33, mean arousal of 4.33, and mean dominance of 4.51.

*3.3 Correlations*

Results of the correlations between rating variables and between tag variables for both conditions are presented in Figure 5. Correlations involving NLP semantic and emotion variables and the complete correlation matrix with the comparison of correlation strength between the two conditions are reported in the Supplementary Material: https://doi.org/10.5281/zenodo.15861012.

Considering rating variables for visual metaphors, we observed strong negative correlations between difficulty and efficacy, as well as between difficulty and arousal ($ps < 0.001$), and a moderate negative correlation between difficulty and artistic quality ($p < 0.001$). This indicates that the more a visual metaphor was rated as difficult to understand, the less it was considered as effective, emotionally activating, and aesthetically pleasing. Also, moderate to strong positive correlations emerged between all variables but difficulty, specifically between efficacy and arousal, efficacy and artistic quality, and arousal and artistic quality ($ps < 0.001$). For literal stimuli, the pattern was similar but weaker. The comparison analysis confirmed that the correlations between difficulty and artistic quality ($p < 0.05$), efficacy and artistic quality, efficacy and difficulty ($ps < 0.01$), difficulty and arousal ($p < 0.05$) were significantly stronger in visual metaphors compared to literal stimuli.

Moving to tag variables for visual metaphors, weak to moderate negative correlations emerged between tag number and word number ($p < 0.001$) and between tag number and non-depicted/depicted ratio ($p = 0.01$). This indicates that the more tags visual metaphors elicited the shorter the tags were and the more they referred to something explicitly visible. Also, a moderate positive correlation emerged between word number and non-depicted/depicted ratio ($p < 0.001$), indicating that the more the words in the tags the more they referred to non-depicted entities. For literal stimuli, only a weak positive correlation emerged between word number and non-depicted/depicted ratio ($p = 0.00$). The comparison analysis of the correlations revealed that the correlation between tag number and word number was significantly stronger in visual metaphors than literal stimuli ($p < 0.01$), but there were no differences in strength for the other correlations.

A number of correlations emerged between semantic and emotion variables on the one hand and rating and tag variables on the other hand. However, no difference in the strength of these correlations between metaphorical and literal items emerged (see Supplementary Material).

To summarize, the main results stemming out of the correlational analysis were the stronger negative associations for visual metaphors compared to literal items between difficulty and all the other variables that indicated communicative benefits (artistic quality, efficacy, and arousal), as well as the stronger positive association efficacy and artistic quality.

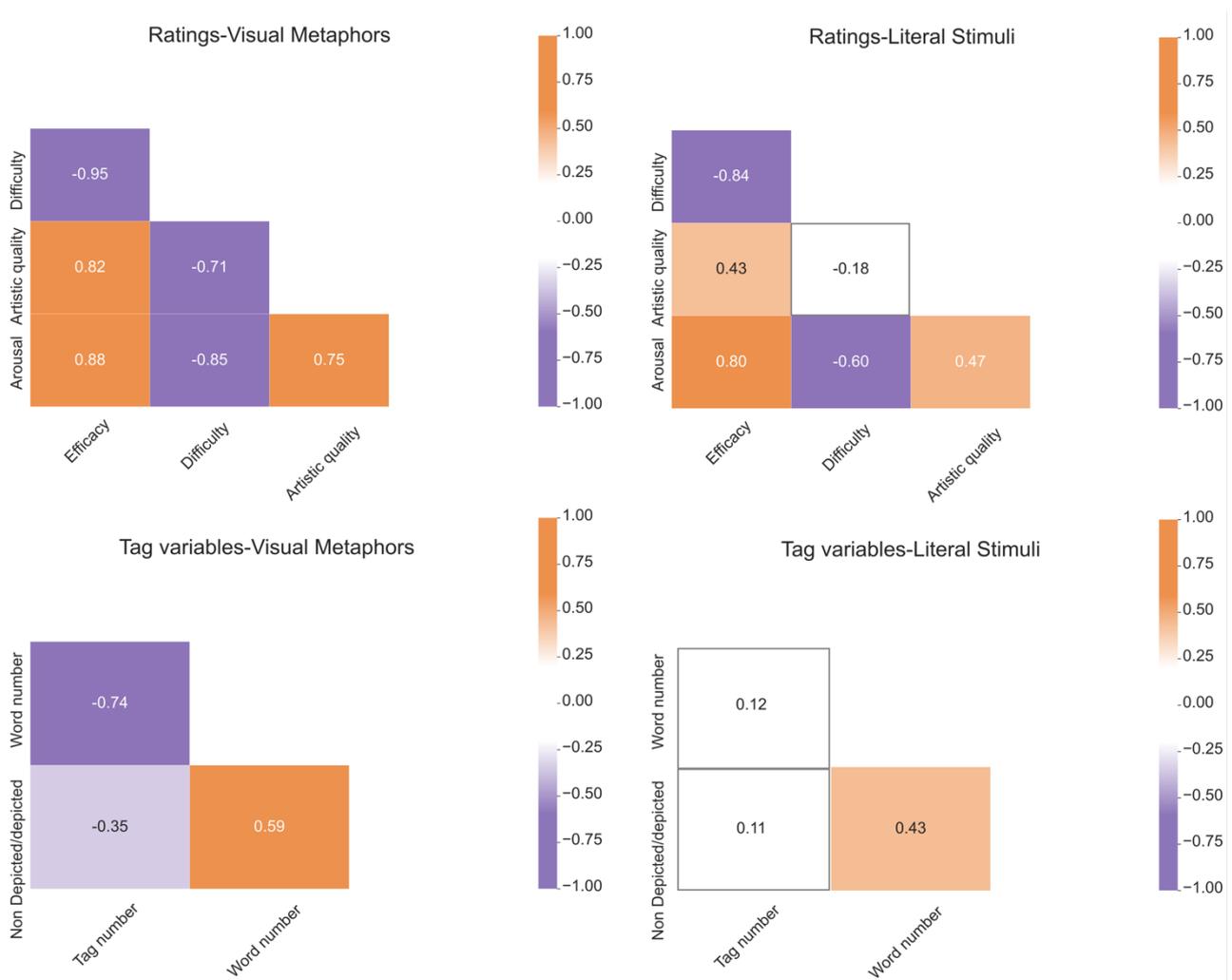

**Figure 5**. Correlograms showing correlations between rating and tag variables. The left column displays correlations for visual metaphors, while the right column represents correlations for literal stimuli. The strength

of the associations is indicated by color, with orange representing positive correlations and purple indicating negative correlations.

For NFC, for visual metaphors, a weak positive correlation was observed with arousal ($p = 0.03$), suggesting that individuals with a higher tendency to seek information experienced a greater perception of emotional activation. For literal stimuli, a weak positive correlation was observed with the number of tags ($p = 0.04$), indicating that individuals with a greater tendency to seek information exhibited a richer tag production. The full correlation matrix is reported in Table 2.

**Table 2.** Comparison of correlations for each variable with NFC measured in visual metaphors vs. literal stimuli.

| Variable 1 | Variable 2 | Visual metaphors | Literal stimuli | Fisher's z | uncorrected p-value | p-value (FDR corrected) |
|---|---|---|---|---|---|---|
| NFC_Total | Tag number | 0.20 | 0.23* | 0.19 | 0.848 | 0.934 |
| NFC_Total | Word number | 0.04 | 0.03 | 0.08 | 0.933 | 0.934 |
| NFC_Total | Difficulty | -0.12 | -0.20 | 0.52 | 0.602 | 0.934 |
| NFC_Total | Efficacy | 0.04 | 0.03 | 0.09 | 0.923 | 0.934 |
| NFC_Total | Artistic Quality | -0.03 | -0.14 | 0.68 | 0.494 | 0.934 |
| NFC_Total | Arousal | 0.25* | 0.15 | 0.58 | 0.561 | 0.934 |
| NFC_Total | Non-depicted/depicted | -0.19 | 0.17 | 2.20 | 0.028 | 0.448 |
| NFC_Total | Concreteness | 0.02 | -0.07 | 0.56 | 0.575 | 0.934 |
| NFC_Total | Imageability | 0.04 | -0.12 | 0.94 | 0.348 | 0.934 |
| NFC_Total | Semantic similarity | -0.20 | -0.12 | 0.44 | 0.662 | 0.934 |
| NFC_Total | Valence | -0.04 | 0.05 | 0.51 | 0.612 | 0.934 |
| NFC_Total | Arousal word | 0.04 | -0.07 | 0.65 | 0.513 | 0.934 |
| NFC_Total | Dominance | -0.03 | 0.07 | 0.60 | 0.546 | 0.934 |

*Note: The first two columns report the two variables tested in the correlation. The third and fourth column represent the values of those correlations in each set of materials, and the remaining columns (Fisher's z, uncorrected p value and p value) describe the statistics testing the differences between the correlations in visual metaphors vs. literal stimuli.*

## 4. Discussion

Despite widespread use of visual metaphors in advertising and environmental communication (Hidalgo-Downing & O'Dowd, 2023), there is very little research investigating the pros and cons of adopting these types of pictures in CC campaigns. So far, studies have primarily focused on the mental representations of CC (Leiserowitz, 2007) or have employed real images of CC from newspapers not being publicly accessible, making it difficult to fully understand their characteristics (O'Neill et al., 2013). The lack of open databases, often limited to literal communication of CC (de Sousa Magalhães et al., 2018; Lehman et al., 2019), might have also limited the scope of the research until now.

The first aim of the present work was to fill this gap by building a database of metaphorical and literal material used in CC visual communication. For this purpose, we constructed the MetaClimage database, which includes fifty metaphors and fifty literal stimuli from various sources about several CC-related phenomena, spanning from deforestation to plastic pollution to glacier melting.

Importantly, for each picture, MetaClimage includes relevant measures useful for the study of multimodal communication (Pérez-Sobrino, 2016), such as rating values of efficacy, difficulty, artistic quality, arousal, as well as a number of measures derived from the tag generation task and indicative of elaboration costs and emotional aspects. In including both literal and visual items, the MetaClimage differs from previous CC visual databases (de Sousa Magalhães et al., 2018; Lehman et al., 2019), as it allows to better account for the variety and complexity of communicative forms (Pérez-Sobrino, 2016). More generally, we contribute with a topic-specific resource to the largely shared efforts to provide readily usable and extensively described material for metaphor studies (Julich-Warpakowski & Perez-Sobrino, 2023; Milenković et al., 2024), up to the recent Figurative Archive (Bressler et al., 2025).

In addition to constructing the dataset, we conducted an initial exploration of the costs and benefits associated with visual metaphors, based on the analysis of the rating and the tag generation task. Capitalizing on previous literature focusing on visual metaphors in advertising (Bambini et al., 2024; Ventalon et al., 2020), we hypothesized that CC visual metaphors would be more cognitively demanding, yet would produce greater benefits at various levels, compared to literal stimuli. Our hypothesis was overall confirmed, in that visual metaphors were associated with a series of measures indicating greater cognitive costs, while being at the same time judged as more artistically pleasing and associated with more positively valenced words. Yet, CC visual metaphors were not considered as more effective and neither as more emotionally impactful in conveying CC messages, which suggests that their advantages do not go to all possible directions, and that choosing to use these expressions requires a careful balance of their costs and benefits.

Starting with the hypothesized overall greater difficulty of visual metaphors compared to literal ones, the supporting findings that we observed are the higher difficulty in the ratings, the increased number of tags generated, and the higher use of tags referring to non-depicted entities and semantically distant concepts. This pattern is in line with prior literature suggesting that visual metaphors are more complex than non-metaphorical pictures, given the presence of a perceptual incongruity that needs to be solved (Bambini et al., 2024; Forceville, 2014). Interestingly, the greater complexity associated with visual metaphors appears to be genuinely conceptual, rather than perceptual, since these items were characterized by a lower entropy, a measure of disorder and unpredictability (Humeau-Heurtier, 2018). While high-entropy images exhibit a wide range of pixel values and features, low-entropy images are more uniform and straightforward from a perceptual point of view. Visual metaphors seem to fall in this latter group, being structurally simpler, albeit more difficult in terms of conceptual elaboration. The absence of verbal anchoring may have further amplified the difficulty of visual metaphors, which is typically attenuated when a verbal explanation is provided (Lagerwerf et al., 2023).

It must be noted that in addition to receiving more tags, visual metaphors elicited a higher number of non-depicted tags compared to literal stimuli. We can speculate that literal stimuli allow for a more direct retrieval of visual elements from memory to derive the intended message, while understanding visual metaphors passes through the initial recognition of the perceptual incongruity, which is intended to be communicative rather than purely descriptive (Schilperoord, 2018) and involves decoding as well as adjusting and inferring from the encoded concepts represented in the images (Yus, 2009). This might result in a greater number of tags and in a greater mentioning of non-depicted entities to derive the message. Results of semantic similarity, which was lower for visual metaphors compared to literal stimuli, can be explained along the same lines. When asked to summarize metaphorical messages, people are drawing upon inferential processes that make them refer to more semantically distant concepts, in line with prior literature evidencing lower semantic similarity for metaphorical terms in visual metaphors (Bambini et al., 2024). The frequent use of novel and distant entities for visual metaphors might also be indicative of a greater abstraction, defined as the process of forming concepts by extracting similarities and general tendencies from direct experience, language, or other concepts (Reilly et al., 2025). Abstraction is reached via a range of different inferential processes (Bolognesi & Vernillo, 2019) and might in turn enhance processes of knowledge transfer (Borghi et al., 2017), contributing to the benefits of visual metaphor to access concepts and knowledge pertaining to CC. Overall, hence, the greater cognitive effort generated by visual metaphors might represent not just a cost but ultimately a benefit in terms of knowledge establishment, with a further range of possible consequences such as, for instance, maintenance in memory, in line with accounts arguing that deeper information processing is associated with higher levels of retention and longer-lasting traces (Craik, 2002).

As for other benefits, visual metaphors stood out for a greater appreciation in terms of aesthetic quality. Indeed, in the tag generation task, participants used words with increased positive valence and dominance when describing visual metaphors compared to literal stimuli. We interpret this result as potentially driven by a greater aesthetic appreciation for metaphors, which might induce a positive affect (McQuarrie & Mick, 2003). These results are in line with a longstanding literature about visual metaphors, typically defined as artful deviations (McQuarrie & Mick, 2003), which, together with the affective involvement, make visual metaphors particularly effective in communication (Phillips, 2003). Despite the greater aesthetic appreciation and the more positive valence at the word level, visual metaphors were judged neither as more arousing nor as more effective than literal stimuli. Even though somewhat unexpected, these results align with previous studies on health-related communication, such as those addressing the damaging effects of tobacco. In anti-tobacco campaigns, images depicting diseased body parts or human tissues were shown to be more arousing and effective compared to visual metaphors, while the latter require careful pre-testing and adaptation to maximize their communicative potential (Wakefield et al., 2013).

Overall, results clearly showed a trade-off between costs and benefits that is unique to visual metaphors. This was also highlighted by the correlation analysis, in particular by the negative correlation between perceived difficulty on the one hand and efficacy, arousal, and artistic quality on the other hand, and by the positive correlation between the latter three. Notably, these correlations were significantly more pronounced for visual metaphors than for literal items, supporting the relevance of the interplay between these variables in the case of metaphors. In this respect, our results confirm prior literature showing that when complexity exceeds a certain threshold, it hampers the comprehension of visual metaphors, vanishing their positive effects (Van Mulken et al., 2014).

We also reported interesting individual differences related to the NFC in the reception of CC messages. Individuals with higher NFC are typically deemed to be more likely to engage in and enjoy cognitive activities (Cacioppo & Petty, 1982) and to find visual metaphors both easier to understand and more enjoyable (Mohanty & Ratneshwar, 2015; Phillips & McQuarrie, 2004; Chang & Yen, 2013). Hence, we hypothesized that individuals with high NFC would report lower difficulty and higher artistic quality for visual metaphors compared to literal images. Results showed that, for visual metaphors, NFC was positively correlated with arousal, suggesting that the higher the NFC, the higher the emotional arousal elicited by the visual metaphor, but no associations emerged for difficulty and artistic quality judgments. There are several reasons that might motivate why our results do not fully align with the previous literature. First, most studies on NFC and visual metaphors used materials from commercial advertising (Mohanty & Ratneshwar, 2015; Kim & Park, 2019), and are therefore scarcely comparable to our study, which focused on CC, a complex topic that engages a specific set of knowledge. Second, the dominance of the most complex types metaphors in the dataset (i.e., fusion and replacement; Phillips & McQuarrie, 2004) may have obscured the facilitatory effect of NFC compared to findings from more heterogenous metaphor type datasets (Mohanty & Ratneshwar, 2015). Also, although individuals with high NFC tend to enjoy cognitively demanding tasks, this might not be captured by the artistic quality judgment, and there may be other relevant individual differences in visual metaphor processing reflecting appreciation, such as participants' taste and propensity (Phillips & McQuarrie, 2004). Indeed, the greater enjoyment for visual metaphors seems in our case to emerge in terms of increased arousal, rather than higher aesthetic appreciation. Although in need of further investigation, our findings are highly relevant in terms of CC communication, as they suggest that the emotional impact of CC-related visual metaphors might depend on the individual's inclination to engage in cognitive activities, rather than being generalized.

It must also be noted that the present study comes with some limitations. The first concerns the MetaClimage database and the possible overrepresentation of plastic-related images. While the database cannot be considered exhaustive with respect to CC topics, it is not unrepresentative of CC communication, given the greater availability of plastic-related pictures online, possibly because plastic pollution was recognized as a major issue by the public earlier than other CC aspects (Windsor et al., 2019). The second limitation relates to the socio-demographic characteristics of the sample. Specifically, participants were relatively young and highly educated individuals, who may be particularly sensitive to CC-related issues (Skeirytė et al., 2022). This factor limits the generalizability of our findings to the broader population. A greater consideration of individual differences beyond NFC, embracing also cultural background, might shed further light on the impact of CC-metaphors, following evidence that indicates that the efficacy of CC communication depends on characteristics such as age, gender, trait empathy and time orientation (Chu, 2022) and considering also that metaphor interpretation varies depending on several aspects, including personality and processing style characteristics (Littlemore, 2019; Battaglini et al., 2025).

Despite these limitations, we believe that a database encompassing visual metaphors and literal stimuli such as MetaClimage can help clarify the complexities of CC communication. Using the database material, we showed that visual metaphors are more difficult, but this produces a chain of benefits that go from a deeper elaboration to a more positive emotional and aesthetic experience, with individual differences related to the individual cognitive profile. While much remains to be clarified in such a complex trade-off of costs and benefits, we believe that MetaClimage can encourage future research to investigate more comprehensively various modes of CC communication (e.g., verbal, visual, multimodal), using both behavioral and neurophysiological measures, to grasp how CC messages can start heating up people's consciousness and promote collective action.

*Data availability statement*

The final MetaClimage database, including results of statistical analysis, is openly available at the following link: https://doi.org/10.5281/zenodo.15861012. A Creative Commons Attribution-NonCommercial-NoDerivatives (CC BY-NC-ND 4.0) license is applied to the database. Note that some images were downsampled to a lower resolution and are free of charge for education and scientific purposes. This research did not receive any specific grant from funding agencies in the public, commercial, or not-for-profit sectors.

*Acknowledgments*

We would like to thank Luigi Cristiano and Teresa Franza from the Technology Transfer and Research Valorization Office at IUSS for their help in retrieving image licenses and assistance in the organization of the MetaClimage database.